\documentclass[final]{anthology-ch} 

\usepackage{booktabs}
\usepackage{graphicx}
\usepackage{multirow}
\usepackage{csvsimple}
\usepackage{longtable}
\usepackage{geometry}
\usepackage{array}
\usepackage[table]{xcolor}
\usepackage{colortbl}
\usepackage{lscape}
\usepackage{subcaption}
\usepackage{wrapfig}

\title{Are You There God? Lightweight Narrative Annotation of Christian Fiction with LMs}

\author[1]{Rebecca M.\ M.\ Hicke}[
  orcid=0009-0006-2074-8376
]

\author[2]{Brian W.\ Haggard}[
  orcid=0000-0001-6199-8642
]

\author[1]{Mia Ferrante}[
  orcid=
]

\author[1]{Rayhan Khanna}[
  orcid=
]

\author[3]{David Mimno}[
  orcid=0000-0001-7510-9404
]

\affiliation{1}{Department of Computer Science, Cornell University, Ithaca, NY, USA}
\affiliation{2}{Department of Sociology, Cornell University, Ithaca, NY, USA}
\affiliation{3}{Department of Information Science, Cornell University, Ithaca, NY, USA}


\pubyear{2025}
\pubvolume{1}
\pagestart{1}
\pageend{1}
\conferencename{Proceedings of Conference XXX}
\conferenceeditors{Editor1 Editor2}
\doi{00000/00000}  

\begin{document}

\maketitle

\begin{abstract}
In addition to its more widely studied cultural movements, American Evangelicalism has a well-developed but less externally visible literary side. Christian Fiction, however, has been little studied, and what scholarly attention there is has focused on the explosively popular \textit{Left Behind} series.
In this work, we use computational tools to provide both a broad topical overview of Christian Fiction as a genre and a more directed exploration of how its authors depict divine acts. 
Working with human annotators, we first developed a codebook for identifying ``acts of God.'' We then adapted the codebook for use by a recent, lightweight LM with the assistance of a much larger model.
The laptop-scale LM is largely capable of matching human annotations, even when the task is subtle and challenging. Using these annotations, we show that significant and meaningful differences exist between divine acts depicted by the \textit{Left Behind} books and Christian Fiction more broadly.
\end{abstract}

\section{Introduction}

The American Evangelical movement has achieved considerable cultural power over the past several decades. 
Accordingly, there has been extensive research on Evangelicalism's historic evolution and efforts to challenge American secularism \cite{braunstein2017prophets, fitzgerald2017evangelicals, gorski2022flag, lieven2012america, smith1998american, whitehead2020taking, du2020jesus}.
However, Evangelical ministry is overwhelmingly male-dominated and, therefore, studies of it center the masculine Evangelical approach to asserting Christian cultural dominance \cite{du2020jesus}.
Christian Fiction, a female-dominated domain \cite{cordero2004production}, has received far less academic study despite having wide readership and billions of dollars in sales \cite{silliman2021reading}.
It formally established itself as a genre in the 1970s, creating a Christian alternative to mainstream secular entertainment \cite{gutjahr2002no}.
Christian Fiction authors occupy a unique position within American Christianity; when they depict God acting in their stories -- speaking to characters, answering prayers, intervening in crises -- they make consequential spiritual and creative choices about divine character and behavior. Existing scholarship on Christian Fiction has been limited and skewed towards the commercially dominant \textit{Left Behind} series \cite{frykholm2004rapture, gutjahr2002no, swirski2014sacrifice, napierala2012strategy, chapman2009tender, dittmer2008geographical, lampert2012left, ashton2019young}. Written by two men in the turn of the 20th century, this apocalyptic series does not represent what Christian Fiction is today: a women-dominated industry of primarily contemporary and historical romance, producing hundreds of titles annually that center women's spiritual lives and everyday faith \cite{neal2006romancing}.

In this paper, we seek to provide a contemporary overview of Christian Fiction using computational methodologies.
We study a corpus of 88 novels: 80 that won or were short-listed for a Christy Award, offered by the Evangelical Christian Publishers Association for excellence in Christian Fiction, and the remaining 8 novels from the \textit{Left Behind} series.\footnote{Four novels from the 12 book series were included in our initial 80-book sample of Christy Award honorees.}
We begin by characterizing the main themes of these works using an established method, statistical topic modeling.
We find that the collection encompasses a wide range of themes, only some of which are distinctively Christian. 
The \textit{Left Behind} books differ significantly from the rest of the collection along several thematic axes.
Moreover, even when the \textit{Left Behind} books are removed from the corpus, themes depicted by male and female authors differ in traditionally gendered ways.
Next, we identify and characterize how authors portray acts of God in their stories.
To do this, we adapt human annotation prompts for use with a small, lightweight LM (Gemma 3n) using a much larger model (GPT-4o). 
Again, we find that the \textit{Left Behind} novels stand out from the rest of the corpus; they feature acts of God significantly more frequently, depict more divine acts that affect groups, and are more likely to characterize those acts as punishing.

Overall, we show that the frequently spotlighted \textit{Left Behind} series is not representative of the general themes of Christian Fiction or its typical portrayal of divine intervention; instead, the authors who dominate the contemporary Christian Fiction landscape portray a loving God who overwhelmingly intercedes on the behalf of individuals. We also find some differences between contemporary male and female authors; namely, women authors focus more on joyful depcitions of affection and less on violence than their male counterparts. Finally, this paper demonstrates that large, contemporary LMs can be used to effectively adapt human annotation prompts for use with smaller LMs, even for highly subtle and complex narrative phenomena.

\section{Background}

\subsection{Evangelicalism + Christian Fiction}

Evangelicalism as an American religio-cultural movement defies simple categorization, with scholars offering a variety of overlapping yet distinct definitions \cite{silliman2021reading, fitzgerald2017evangelicals, butler2024white, gorski2020american}. Despite this debate, Evangelicals are generally distinguished from the broader Christian community by several core beliefs: upholding the Bible as the ultimate spiritual authority that should be read and interpreted literally, experiencing a born-again conversion or moment of salvation, celebrating the centrality of Christ's atonement, and actively working to spread their faith \cite{du2020jesus, neal2006romancing, wong2018immigrants}. Christian Fiction specifically gained momentum in the 1970s as American Evangelicalism emerged as a mainstream cultural movement responding to social upheaval around race, gender, and sexuality \cite{napierala2012strategy}. Rather than shunning mainstream culture as earlier Christian Fundamentalists had done, Evangelicals increasingly encouraged the creation of alternatives from secular consumption, contributing to the growth of the genre.

Twentieth-century Evangelicalism intensified patriarchy and women's exclusion from formal ministry even as broader American culture moved toward equality \cite{du2020jesus, barr2021making}. However, Christian Fiction represents one of the few spaces within the Evangelical movement where women can exercise ministerial authority. Authors within this genre describe their writing as divine calling, positioning themselves as spiritual vessels using narrative to facilitate readers' encounters with their faith. Following pioneers like Janette Oke who established the legitimacy of Christian romance, contemporary authors understand their creative process as fundamentally spiritual, involving prayer and calling for God's guidance when developing plot and character. Readers engage with these texts as devotional practice, using fiction to maintain faith and understand God's character \cite{frykholm2004rapture, paradis2011typological, neal2006romancing, silliman2021reading}. This creates ``a ministry by, for, and about women'' \cite[p.~108]{neal2006romancing} operating outside institutional structures that formally restrict women's spiritual leadership.

Examining how authors portray God's actions allows us to directly study their perceptions of God. Authors make consequential choices about divine behavior within their narratives: how God speaks, what motivates divine intervention, how God responds to suffering, and more \cite{chapman2009tender, douglas2022silence}. These choices reflect authors' theological convictions; authors would not write God acting in ways they consider blasphemous or contrary to scripture. When an author depicts God comforting an individual, calling a nation to repentance, or remaining silent during tragedy, they therefore communicate their understanding of God to readers. These narrative patterns thus reveal how Evangelical women conceptualize and teach about God's relationship to humanity, offering insight into women's religious expression.

Within Christian Fiction, the \textit{Left Behind} series by Tim LaHaye and Jerry B.\ Jenkins in particular has come to dominate both the market and the scholarly understanding of the genre. 
\textit{Left Behind}'s success cannot be understated. It sold over 80 million copies and thrust the once niche Christian Fiction industry into the mainstream \cite{domonoske2016tim}. The 12 book series depicts a prophetic future for the world informed by LaHaye's literalist interpretation of the Book of Revelation \cite{swirski2014sacrifice}. The series explores how the End Time descends upon the world as the Antichrist battles for control over the planet and is set within a soon approaching future, incorporating real locations, governments, and political parties. The series generated substantial academic attention that still constitutes the bulk of research on Christian Fiction. This includes discussion of the books’ perspective on Christian faith \cite{gutjahr2002no, gribben2004rapture}, race \cite{mcalister2003prophecy}, gender \cite{chapman2009tender}, and politics \cite{napierala2012strategy, dittmer2008geographical}. Yet the singular focus of this scholarship presents a significant problem; \textit{Left Behind} is not representative of Christian Fiction. The series is increasingly outdated --- the final book in the \textit{Left Behind} series was released over 20 years ago. More importantly, \textit{Left Behind} was co-authored by a prominent figure in Evangelical political movements and its apocalyptic narrative contains patriarchal themes potentially unrepresentative of the novels that define contemporary Christian Fiction \cite{chapman2009tender, du2020jesus}. While some scholarship has addressed more general trends in Christian Fiction \cite{silliman2021reading} or women-dominated genres \cite{neal2006romancing, barrett2003higher}, these studies remain limited in their examination of the broader Christian Fiction community. These studies also focus on a very small number of texts, limiting their generalizability.

\subsection{Narrative Analysis with LMs}

Language models are increasingly being used for narrative analysis tasks. Many studies have employed a variety of prompting strategies with large, generative LMs for narrative analysis tasks involving character relationships, roles, and traits \cite{laato2024extracting, yuan-etal-2024-evaluating, jaipersaud2024show, sancheti2025tracking, brei2025classifying, anglin2025scaling}; narrative flow and style \cite{hicke2024says, sunny2025stories}; and much more \cite{jang2024evaluating, kampen2025llm, shen2024heart, jenner2025using, michel2025evaluating, subbiah2024reading}.
These studies often rely on human annotation and inter-annotator agreement to assess how closely LM judgments align with human readers \cite{paragini2022roots, jang2024evaluating}. While some studies find LM performance to exceed baselines and show great promise for narrative understanding \cite{jaipersaud2024show}, others find LLMs struggle with subtext and ambiguity. For example, while LMs can often produce plausible narrative summaries or identify who said what in a dialogue, they still frequently struggle with deeper interpretive tasks that involve implicit reasoning or contradictory cues \cite{jang2024evaluating}.

Our work builds on this growing area of research by applying LLMs to a deeply nuanced interpretive task: identifying instances of divine intervention in Christian Fiction. This task extends beyond plot analysis, requiring theological insight and cultural context. We know of no existing research which uses LLMs to study narrative and faith in Christian Fiction.

\subsection{Automatic Prompt Generation}

Researchers have recently explored automatic prompt generation with LLMs.  \textcite{honovich2023instruction} evaluate models ability to produce task instructions by using generated instructions as model prompts. Several studies have used LLMs to produce and improve natural language prompts \cite{zhou2023large, wang2023self, fernando2024promptbreeder, ye2024prompt, deng2023rephrase}.
However, no work we are aware of uses human and model collaboration to iteratively refine narrative analysis prompts for smaller LMs. 

\section{Data}

\begin{table}[ht!]
\centering
\begin{tabular}{lcc}
\toprule
 & \textbf{Christy Award Corpus} & \textbf{Our Corpus} \\
\midrule
\textbf{\# Texts} & 647 & 80 \\
\textbf{\# Publishers} & 75 & 12 \\
\textbf{\# Unique Authors} & 335 & 71\\
\textbf{\% Female Authors}* & 76\% & 74\% \\
\textbf{\% Male Authors}* & 23\% & 26\%\\
\textbf{\# Award Categories} & 21 & 18\\
\textbf{\% Multi-Award Nominees} & $4.5\%$ & $15\%$\\
\bottomrule
\end{tabular}
\caption{Descriptive comparisons of our subcorpus (right) vs.\ the entire list of Christy Award honorees (left). Statistics marked with a * exclude novels co-authored by a male and female author.}
\label{tab:corpusComp}
\end{table}

Our corpus includes 80 Christian Fiction novels by 71 authors published between 2000 and 2023.
These books are sampled from a comprehensive list of Christy Award book winners and finalists published between 2000 and 2024.
The Christy Award was established by the Evangelical Christian Publishers Association in 1999 to celebrate faith-based novels' impact on contemporary culture.
Books honored by the Christy Award have been acknowledged as both high quality and culturally significant within the Evangelical community. 
By studying these books we ensure our sample represents the most influential and respected voices in Christian Fiction.

The full corpus of Christy Award honorees comprises 647 titles from 75 publishers. Over 25 years, awards have been given in 22 distinct categories ranging from `First Novel' to `Suspense' and `North American Historical.' Approximately three-quarters of Christy Award honorees were authored only by women.\footnote{Some books were co-authored by male and female authors and are excluded from this count. A breakdown of the proportion of honorees written women for each award category can be found in Appendix \ref{appdx:catByGender}.} In our dataset, we include all Book of the Year awardees from 2014 to 2023. The remaining 70 books we sample from the 17 other award categories of interest,\footnote{The three excluded categories are the Amplify Award (only given in 2022 and 2023), Lits (only given in 2007 and 2008), and Short Form (which includes novellas and short stories, but not novels).} prioritizing ebook accessibility. Our corpus includes novels from 12 publishers, five of which account for over 61\% of Christy Award honorees.\footnote{These publishers are: Bethany House Publishing, Revell / Baker Publishing Group, Thomas Nelson, Tyndale House Publishers, and WaterBrook Multnomah Publishing Group} As in the collection of all honoreees, three-quarters of the books in our subset were authored by women. The number of books we sampled from each award category can be found in Appendix \ref{appdx:corpusByCat}. 15\% of books in our corpus won multiple Christy awards, compared to 4.5\% in the entire dataset, because all Book of the Year awardees by definition won another award in the same year. An overview comparison of our subsample to the entire set of Christy Award honorees can be found in Table \ref{tab:corpusComp}.

To the set of 80 Christy Award honorees, we add the remaining eight novels from the main \textit{Left Behind} series (12 books total) to facilitate comparison between this series and the Christian Fiction genre more broadly. The additional books were published between 1995 and 2004 by Tyndale House Publishers. Details on all 88 books in our corpus can be found in Appendix \ref{appdx:corpus}.

\section{Methods}

Our goal is to observe and quantify the thematic contents of our collection of Christian Fiction, particularly the characterization of the divine.
We approach this question from both a broad, unsupervised perspective and from a more specific, targeted perspective.
These goals require different methodologies.

\subsection{Topic Modeling}

For the broad thematic perspective, we find that a standard LDA topic model \cite{blei2003latent} is both effective and well-established. 
We used the Mallet toolkit \cite{mccallum2002mallet} with hyperparameter optimization for $\alpha$ and $\beta$.
We chose the granularity of topics ($K=65$) by inspection for a contextually appropriate balance between comprehensiveness and specificity.

Because we are working with long-form fiction, we preprocess the novels in two ways.
First, we divide each novel into 300 word segments.
This scale results in 29,000 ``documents'' rather than 88, which provides more statistical support for thematic analysis.
Second, we use the Authorless Topic Models method \cite{thompson2018authorless} to reduce the impact of novel-specific character names and settings and emphasize themes that occur across multiple works.
This step is similar to a contextual stopword list, but rather than fully removing words it stochastically reduces the frequency of overrepresented words in specific novels.
Together with a customized stopword list, these preprocessing steps reduce the number of word tokens from roughly 8.8 million to 2.8 million.

\subsection{Identifying Acts of God}

For a more focused analysis, we sought to identify and characterize acts of God in our corpus.
Though the books are fictitious, guidelines for publishing Christian Fiction \cite{cordero2004production} and authors' unwillingness to inaccurately portray God lead to stories that represent the authors' religious beliefs.
This contributes to the blurring of the line between sacred and fictional aspects of Christian Fiction \cite{gutjahr2002no, swirski2014sacrifice, frykholm2004rapture, paradis2011typological}.
Identifying and classifying these acts of God required both significant interdisciplinary scholarship and sophisticated contemporary language models.

\subsubsection{Codebook Creation}

For annotation, we split the books in our corpus into $\leq$500 word passages.
500 words constitute about two average novel pages; using passages of this length allows us to provide annotators with contextually rich passages while staying within the high-performing context lengths of smaller LMs.
We preserve chapter and section breaks in the novels to further increase the coherency of passages.
This results in 20,440 unique passages. On average, there are 232.27 passages per volume, with a minimum of 29 passages\footnote{\textit{Candle in the Darkness} by Lynn Austin} and a maximum of 473 passages.\footnote{\textit{The Indwelling} by Jerry B.\ Jenkins and Tim LaHaye} The passages are between 2 and 500 words long, with an average length of 430.80 words.

The codebook and annotations were created by a team of ten: the lead authors with degrees in literature, sociology, and computer science and eight undergraduates with a variety of religious and academic backgrounds. Initially, we expected the coding task to be straightforward; the undergraduates were instructed to code a passage as ``Yes'' if God explicitly acted and ``No'' otherwise. However, the Krippendorff's $\alpha$ scores \cite{krippendorff1970estimating} for the first two annotations rounds were very low (0.36 and 0.48). Perhaps unsurprisingly, we found that God’s presence in Christian Fiction narratives is multidimensional and often challenging to concretely identify. We also found that students' familiarity with and relationship to Christian theology dramatically changed their interpretations of what was considered an ``act'' of God. To develop a unified and concrete understanding of Evangelical depictions of God, the team spent two semesters meeting weekly. Students were assigned relevant literature on American Christianity, Evangelicalism, and Christian Fiction to establish a common language and knowledge of the topic. Each week, students were tasked with coding a set of passages and asked to present on passages they found challenging. 

Student feedback and extensive discussions among the entire team resulted in several key changes to the codebook. First, we decided that it was important to code from an Evangelical perspective. This meant that a passage was labeled ``Yes'' whenever an action was attributed to God, however mundane. Therefore, we included actions ranging from thanking God for providing a delicious meal to God sending apocalyptic plagues. Second, the adoption of the Evangelical perspective meant that the Bible was considered the literal word of God and Bible quotes or stories describing God's actions were labeled ``Yes.'' Finally, we introduced a ``Maybe'' label to cover edge cases. These changes resulted in an increase in inter-rater reliability to $\alpha\geq0.65$. Given the difficulty of task and the large number of annotators, we considered this sufficient.

We chose the $\leq$500 word segmentation method because it was unambiguous and could be implemented quickly and accurately, but there are both technical and literary complications. 
In some cases, acts of God would fall across the segmentation boundaries, making them difficult to accurately identify.
There were also potential acts of God that would later be revealed as intentional misdirection by the author.
Since annotators only used passage-level context, these acts would still be labeled ``Yes.''
Finally, narrative-specific references to God's actions may have been missed because the coders were not familiar with the entire texts.
We do not believe, however, that these issues were common enough to threaten the validity of our results.

Overall, the annotation process resulted in 1,951 unique coded passages from 67 novels. Of these passages, 1,679 (86.06\%) were labeled as \textit{not} containing acts of God, 191 (9.79\%) were labeled as containing acts of God, and 81 (4.15\%) were labeled as possibly containing acts of God. All passages were initially coded by at least two separate undergraduate annotators. All annotator disagreements were discussed and resolved by the lead authors.

\subsubsection{Annotation with LLMs}

In order to identify acts of God in the entire corpus, we transform the instructions designed for human annotators into two zero-shot prompts, which we then apply using Gemma 3n (e4b) quantized to 4 bits \cite{gonzalez2025gemma3n}.\footnote{The model was accessed via \href{https://ollama.com}{Ollama}.}
We selected Gemma 3n (e4b) for two reasons.
First, we intend this method to be accessible to researchers without significant computer resources; when quantized, the Gemma 3n model can be run on a personal laptop using systems like Ollama for no additional cost.
All experiments for this paper are run on a desktop AMD Ryzen 7 CPU, RTX 4070 GPU, and 32 GB RAM.\footnote{Although a more powerful device was used to speed up inference, the experiments also run successfully on a 2022 Macbook Pro with an M2 chip.}
Second, because this project studies copyright-protected data, we avoid using any models that require upload to a cloud system or that may ingest novel excerpts as training data. Temperature was set to 0 to optimize the model's performance for accuracy

\begin{figure*}
    \centering

    \begin{subfigure}[t]{0.48\textwidth}
        \centering
        \vspace*{0pt}
        \includegraphics[width=\linewidth]{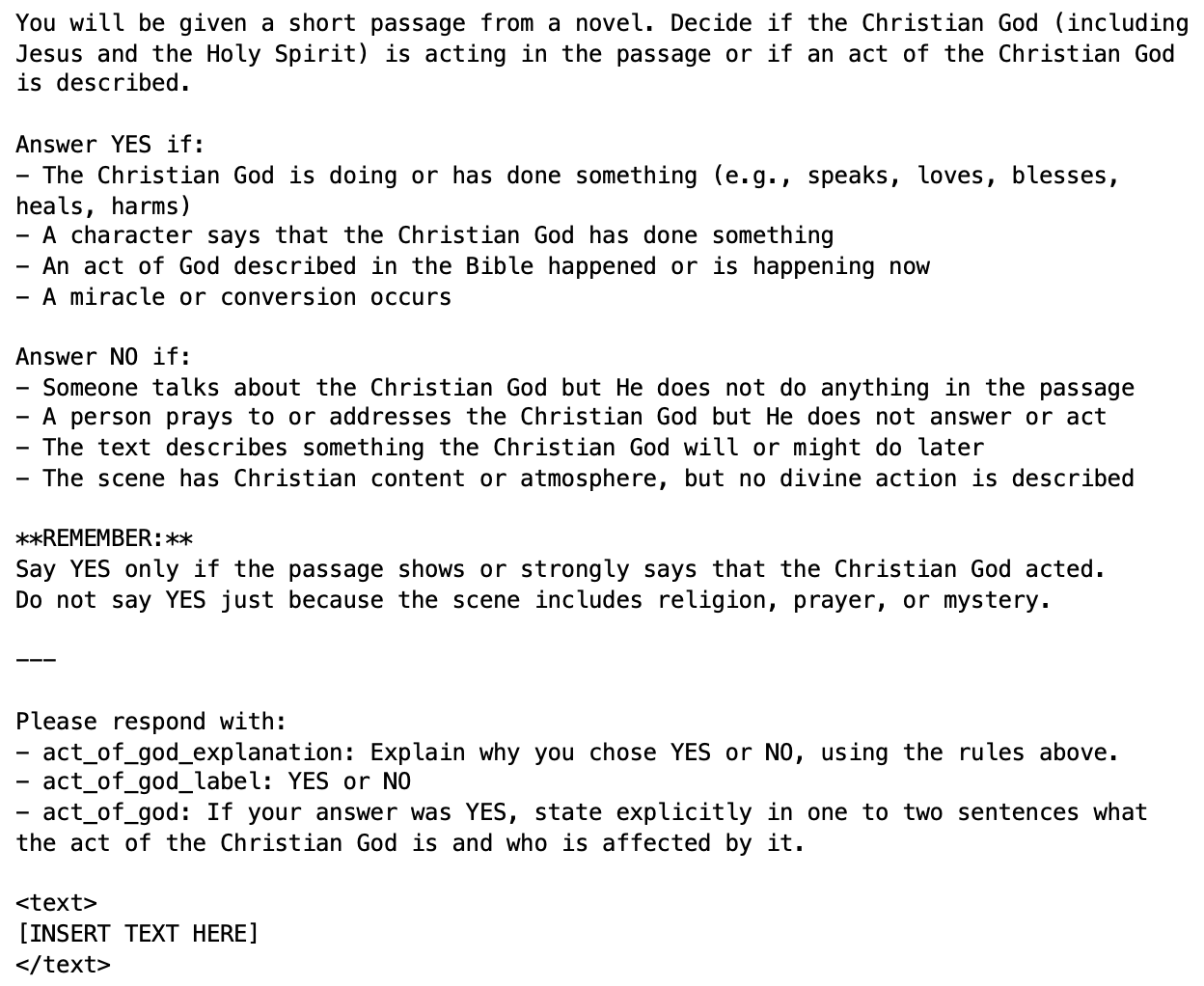}
    \end{subfigure}
    \hfill
    \begin{subfigure}[t]{0.48\textwidth}
        \centering
        \vspace*{0pt}
        \includegraphics[width=\linewidth]{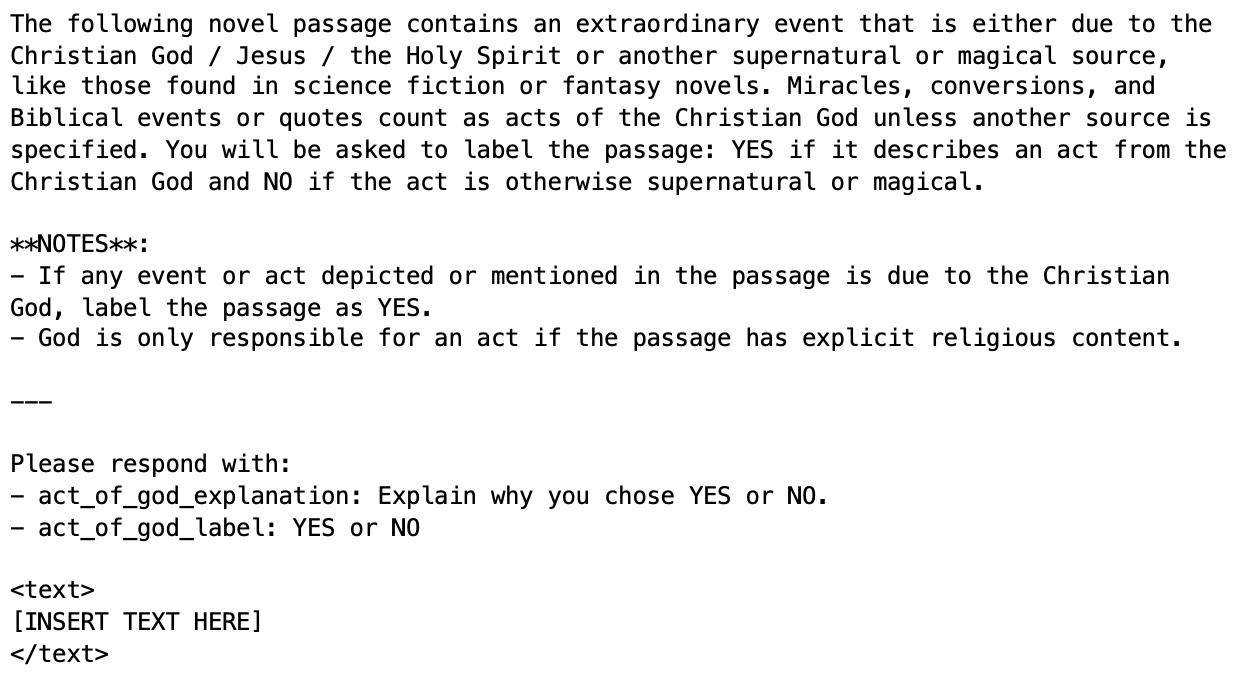}
    \end{subfigure}

    \caption{The two zero-shot prompts used to identify acts of God.}
    \label{fig:actOfGodPrompts}
\end{figure*}

To convert the original annotator instructions (Appendix \ref{appdx:humanActPrompt}) into a set of zero-shot prompts optimized for Gemma 3n, we first simplified the language and syntax, removed redundant instructions, and specified output format. We asked the model to respond with a) an explanation of which label it will choose, b) the label itself, and c) a description of the act of God and who it affects. We used JSON-formatted structured outputs to ensure that the model would provide answers for all three parts and only respond with a recognized label. Next, we used GPT-4o \cite{openai2024gpt4o} interactively to make iterative refinements to the prompt by instructing it to optimize changes to the prompt for a small model, specifically Gemma 3n. GPT-4o further simplified the language in our prompt and introduced additional formatting to add structure.

After each round of changes to the prompt, we tested it on a subset of the evaluation corpus (50 random excerpts labeled ``Yes,'' 50 labeled ``Maybe,'' and 50 labeled ``No''). Based on observed trends in the model responses, the lead authors and GPT-4o made further adjustments to the prompt. In a few cases, we passed specific passages along with the Gemma 3n label explanations to GPT-4o as examples of common mistakes. During this iterative process, the descriptions for each label were heavily edited; instructions were added to catch common mistakes and notes addressing uncommon edge cases were removed to avoid overwhelming the model. In addition, we chose to remove the ``Maybe'' label from the prompt because the distinctions between ``Yes'' and ``Maybe'' were very challenging for Gemma 3n. All ``Maybe'' labels in the evaluation dataset were converted to ``Yes.''

The model also repeatedly struggled to distinguish between acts of God and other supernatural or magical events, like those depicted in science fiction or fantasy novels. Our attempts to clarify this distinction in the original prompt led to decreased model performance for other passages. Therefore, we introduced a second prompt to the classification pipeline that asked the model to clarify between acts of God and supernatural events. All passages labeled ``Yes'' by the original classifier were fed to the second prompt; passages only received the final ``Yes'' label if both prompts labeled them ``Yes.'' The second prompt went through the same iterative editing process by GPT-4o and the research team. The final versions of each prompt are available in Figure \ref{fig:actOfGodPrompts}.

\begin{wraptable}{r}{0.5\textwidth}
    \small
    \centering
    \begin{tabular}{lccc}
        \toprule
        & \textbf{Recall} & \textbf{Precision} & \textbf{F1} \\
        \midrule
        \textbf{Yes} & 0.84 & 0.52 & 0.64 \\
        \textbf{No} & 0.87 & 0.97 & 0.92 \\
        \textbf{Overall} & - & - & 0.87\\
        \bottomrule
    \end{tabular}
    \caption{Gemma 3n's performance on the evaluation dataset using the prompts in Figure \ref{fig:actOfGodPrompts}.}
    \label{tab:modelResults}
\end{wraptable}

The classification pipeline achieved an overall F1 score of 0.87 on the entire dataset (Table \ref{tab:modelResults}). We consider this an impressive result for such a subtle and complex task. However, although the recall for both labels is high, we find that the model continued to over-label passages as ``Yes.'' Specifically, it often labeled passages in which a character prayed or otherwise spoke to God as ``Yes,'' although the passage did not include any implied or explicit \textit{response} by God. For an annotation task as nuanced as this, a small LM the size of Gemma 3n is still limited in its capacity. Despite this, we find that iterative editing by the lead authors and a large, powerful LM converted human annotator instructions into a successful prompt for a smaller, accessible LM that allowed for nuanced analysis on a previously infeasible scope.

\subsection{Characterizing Acts of God}

To further characterize the acts of God identified by the classification system described above, we designed two further prompts for Gemma 3n. These prompts seek to identify who is affected by each act of God (individuals or groups) and whether God's action is loving, punishing, both, or neutral. Again, we begin with prompts written for human annotators and transform them for use with Gemma 3n using GPT-4o. Both prompts take as input the descriptions of God's acts output by the primary classifier. Final versions of both prompts can be seen in Appendix \ref{appdx:characPrompts}.

To evaluate the efficacy of these prompts, the lead authors reviewed their responses to the first 100 passages in the evaluation dataset labeled as ``Yes'' by both the original classifiers and human annotators. We found the annotators agreed with 81\% of the labels identifying who an act affected and 88\% of the labels characterizing acts. However, the labels ``Individual'' and ``Loving'' were over-represented in the annotated set (likely because they are over-represented in the entire corpus) and, moreover, were over-predicted by the model. Despite this, we determined that the classifiers provided a strong enough signal to be used for further analysis.

\section{Characterizing Christian Fiction}

We focus on two primary trends in our topic modeling results: expression of faith and portrayal of gender.\footnote{A full list of topics and associated keywords can be found in Appendix \ref{appdx:topicLabels}.} We identify four faith-related topics: Confessional Prayer (\#13: \textit{believe, father, please, forgive}), Glorifying God (\#43: \textit{God, heaven, salvation, amen}), Apocalyptic Faith (\#48: \textit{Israel, tribulation, death, messiah}), and Congregation Worship (\#63: \textit{church, Sunday, music, sermon}). Surprisingly, these topics are comparatively rare despite the overall importance of faith in the genre (Appendix \ref{appdx:topicProm}). Glorifying God and Apocalyptic Faith are highly positively correlated (Pearson's R: 0.84, $p<10^{-21}$),\footnote{All $p$-values reported have had Bonferroni correction applied using the \texttt{statsmodels} package to account for multiple similar statistical tests being run on the same dataset.} but no significant correlations exist between the other faith categories; this suggests that distinct faith styles are employed by authors.


We also identify five topics tied to complementarianism, the theological argument that men and women were created to serve distinct roles, with men taking on leadership positions in the household and wider society while women primarily serve the family unit. These topics are Food and Cooking (\#1: \textit{food, kitchen, meal, hungry}), Family Members (\#4: \textit{family, father, children, parents}), Family Relationships (\#8: \textit{mom, dad, father, kids}), Motherhood and Pregnancy (\#27: \textit{baby, mother, pregnant, birth}), and Cleaning the Home (\#34: \textit{clean, bathroom, towel, wet}). Prevalence of the Cooking, Family Relationships, and Cleaning the Home topics are all significantly positively correlated as are Family Members and Motherhood and Pregnancy. The relationships between these topics suggests that Christian Fiction depicts an idealized lifestyle for Evangelical women that emphasizes domesticity \cite{du2020jesus}.


These topic groupings provide evidence that the \textit{Left Behind} series differs considerably from Christian Fiction more broadly. Although there is no significant difference between the prevalence of the Confessional Prayer and Congregation Worship topics in the \textit{Left Behind} books and the broader corpus, both the Glorifying God (4.51 vs.\ 1.38, $p<10^{-17}$) and Apocalyptic Faith (5.39 vs.\ 0.41, $p<10^{-37}$) topics were significantly more dominant in the \textit{Left Behind} books.\footnote{Independent t-tests for difference in means were performed using \texttt{sklearn}} The \textit{Left Behind} series also features Cooking and Cleaning the Home at significantly lower frequencies than the rest of the corpus.\footnote{\#1: 0.45 vs.\ 1.75, ($p<0.01$), \#34: 0.37 vs.\ 0.77 ($p<0.05$))} It is unclear whether these differences stem from \textit{Left Behind}’s apocalyptic setting or gendered preferences in writing. 

To pursue this question, we examine whether the prevalence of these topics differs significantly in books written by female and male authors. To do this, we exclude all \textit{Left Behind} novels and books co-written by a male and female author from the dataset. We find no significant differences between topic prominence between male and female authors for any of the four faith-focused topics or any of the five topics tied to complementarianism. However, we do see that women authors depict Joyful Affection (\#2: \textit{love, smile, kiss, joy}) more on average (1.02 vs.\ 1.99, $p<0.05$) and Violence (\#23: \textit{blood, body, gun, dead}) less on average (5.30 vs.\ 3.16, $p<0.05$) than male authors, which aligns with standards of Evangelical femininity.

Overall, we identify three primary patterns in the topic model: the \textit{Left Behind} series approaches faith in a distinct way, this difference in religious portrayal is not due to gender --- male and female authors portray faith to similar extents, but an author’s gender does shift the prevalence of some gendered topics. This makes clear the divide between the intensely religious themes of the \textit{Left Behind} series and more contemporary Christian Fiction.

\section{Identifying Acts of God}

\begin{figure*}
    \centering
    \includegraphics[width=.65\linewidth]{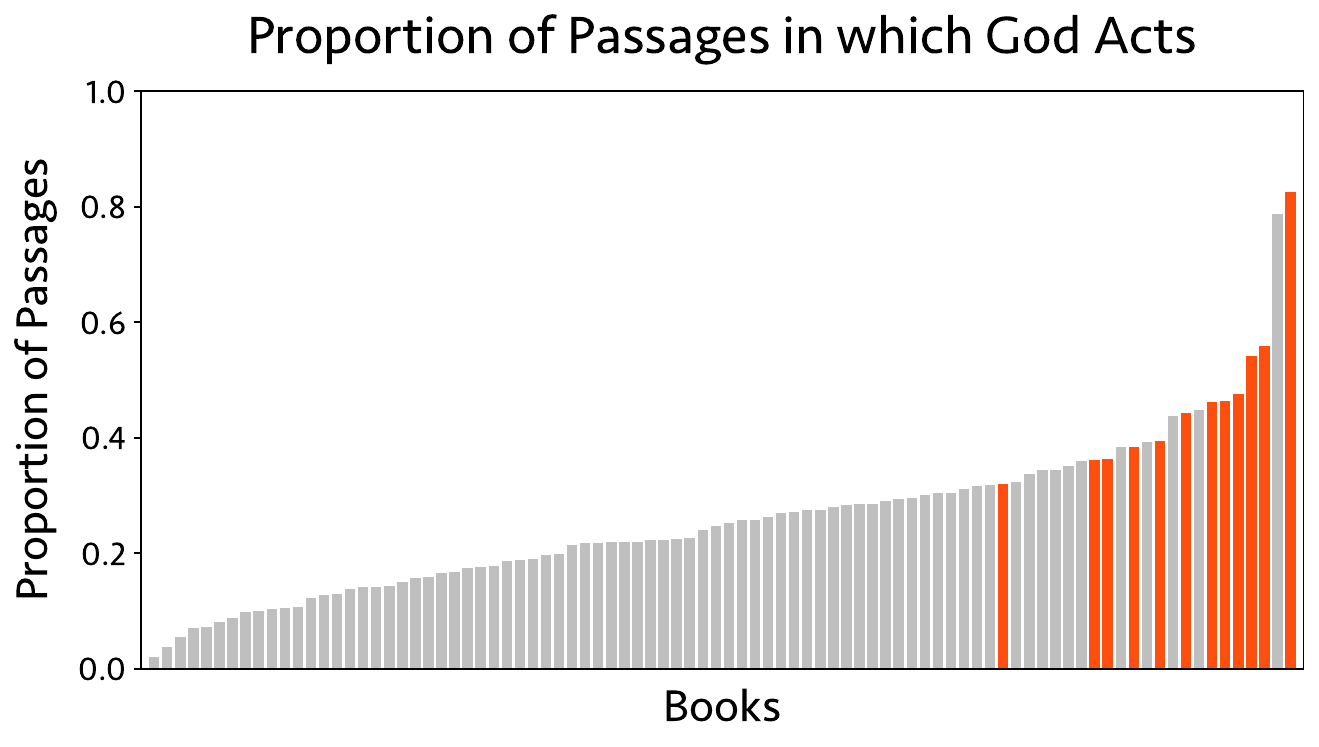}
    \caption{Novels sorted by proportion of passages  labeled as containing acts of God by our classification pipeline. The \textit{Left Behind} novels, shown in orange, have some of the highest proportions.}
    \label{fig:propActs}
\end{figure*}

Of all the passages in our corpus, 24.9\%  or 5,086 are labeled as containing acts of God. On average, 26.1\% of passages from each novel contain an act of God; however, this value ranges from 2.1\%\footnote{\textit{Auralia's Colors} by Jeffrey Overstreet} to 82.6\%\footnote{\textit{Glorious Appearing} by Jerry B.\ Jenkings and Tim LaHaye} across all novels (Figure \ref{fig:propActs}). This demonstrates that, although these books are positioned in an explicitly religious subgenre often assumed to contain religious indoctrination, there is actually wide variation in how frequently authors portray acts of God. 

We find that two of the four faith topics --- Glorifying God (Pearson's R: 0.88, $p<10^{-28}$), and Apocalyptic Faith (Pearson's R: 0.71, $p<10^{13}$) --- significantly correlate with the frequency of divine action. When the \textit{Left Behind} novels are removed from the dataset, these correlations hold\footnote{\#43 -- 0.82 ($p<10^{-18}$), \#48 -- 0.55 ($p<10^{-5}$)} and a significant relationship between the frequency of divine action and Confessional Prayer emerges.\footnote{0.38, $p<0.05$} However, no such significant relationship exists for Congregation Worship. This highlights that explicit actions of God occur more frequently in books which emphasize individualized expressions of faith, which aligns with American Evangelicalism's hyper-individualistic tendencies. 

\begin{figure*}
    \centering
    \includegraphics[width=.65\linewidth]{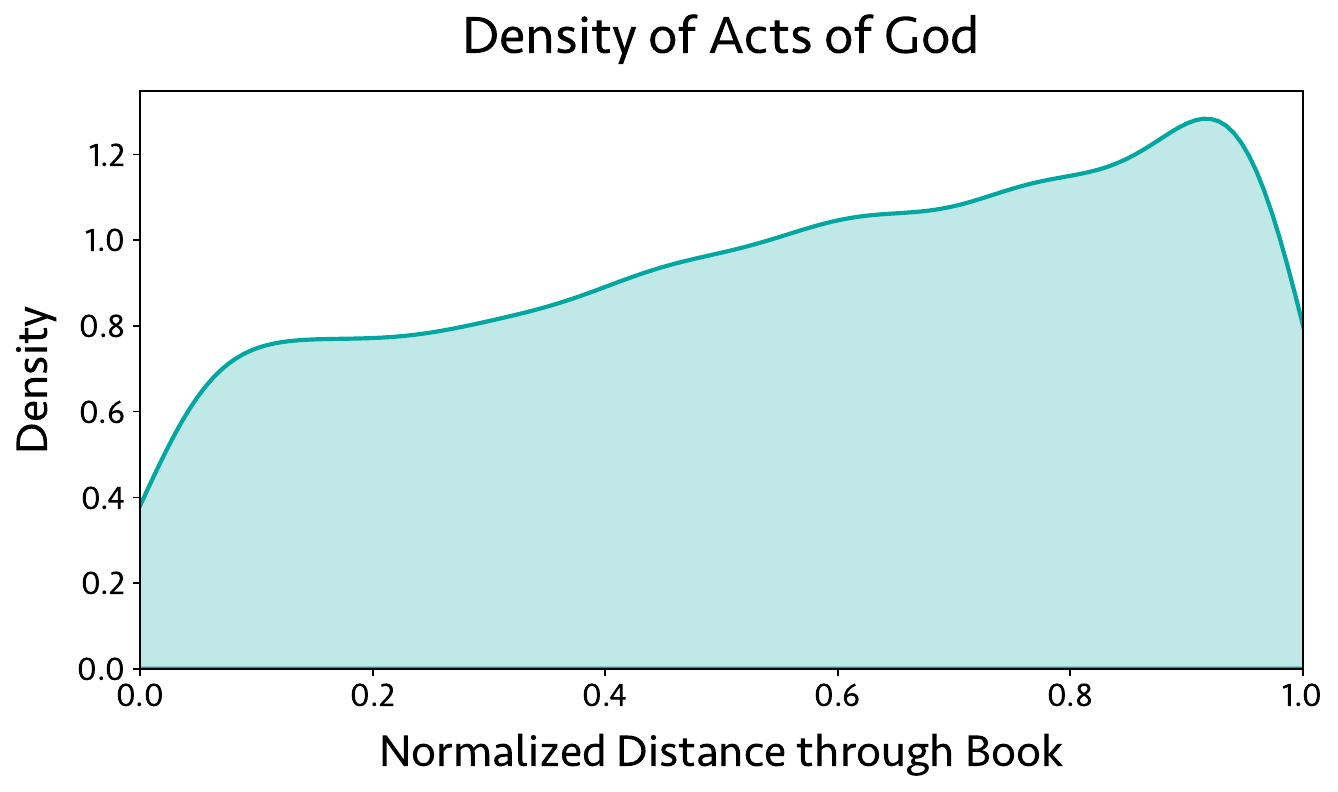}
    \caption{The frequency of acts of God over normalized novel  progression. Acts can occur at any point, but are more frequent later in novels.}
    \label{fig:actDensity}
\end{figure*}

We also see that acts of God are more densely clustered later in novels (Figure \ref{fig:actDensity}). 
On average, acts of God appear 56.2\% of the way through a plot.
There are two possible explanations for this.
First, acts of God may simply serve as important narrative events, which frequently occur towards the end of novels.
Second, Christian Fiction authors may wish to defer ``religious talk'' to a point where readers are invested in the characters or plot, so as to avoid putting off secular readers.
This approach serves to increase readership, but also serves a spiritual goal of introducing non-Christian readers to Christian values and viewpoints.

Finally, we see that there is no significant difference between the frequency with which male and female authors depict acts of God. However, a significantly higher proportion of passages from the \textit{Left Behind} novels depict acts of God on average (46.6\% vs.\ 22.85\%, $p<10^{-9}$);\footnote{All t-tests for difference in means are again performed with \texttt{sklearn} with Bonferroni correction with \texttt{statsmodels}.} this is clearly visible in Figure \ref{fig:propActs}. \textit{Left Behind} explicitly sets out to narrate events from the Book of Revelation and in doing so frequently portrays God as a core actor. In contrast, most books invoke God's intervention more sparingly. Notably, the novel featuring acts of God with the second-highest frequency (\textit{Unashamed} by Francine Rivers) is not from the \textit{Left Behind} series but \textit{is} a retelling of the story of Rahab from the Bible. We would expect books explicitly retelling stories from the Bible to feature more acts of God; their high ranking by our classifier therefore reinforces our confidence in its accuracy.

\section{Characterizing Acts of God}

\begin{figure*}[t]
    \centering
    \includegraphics[width=.65\linewidth]{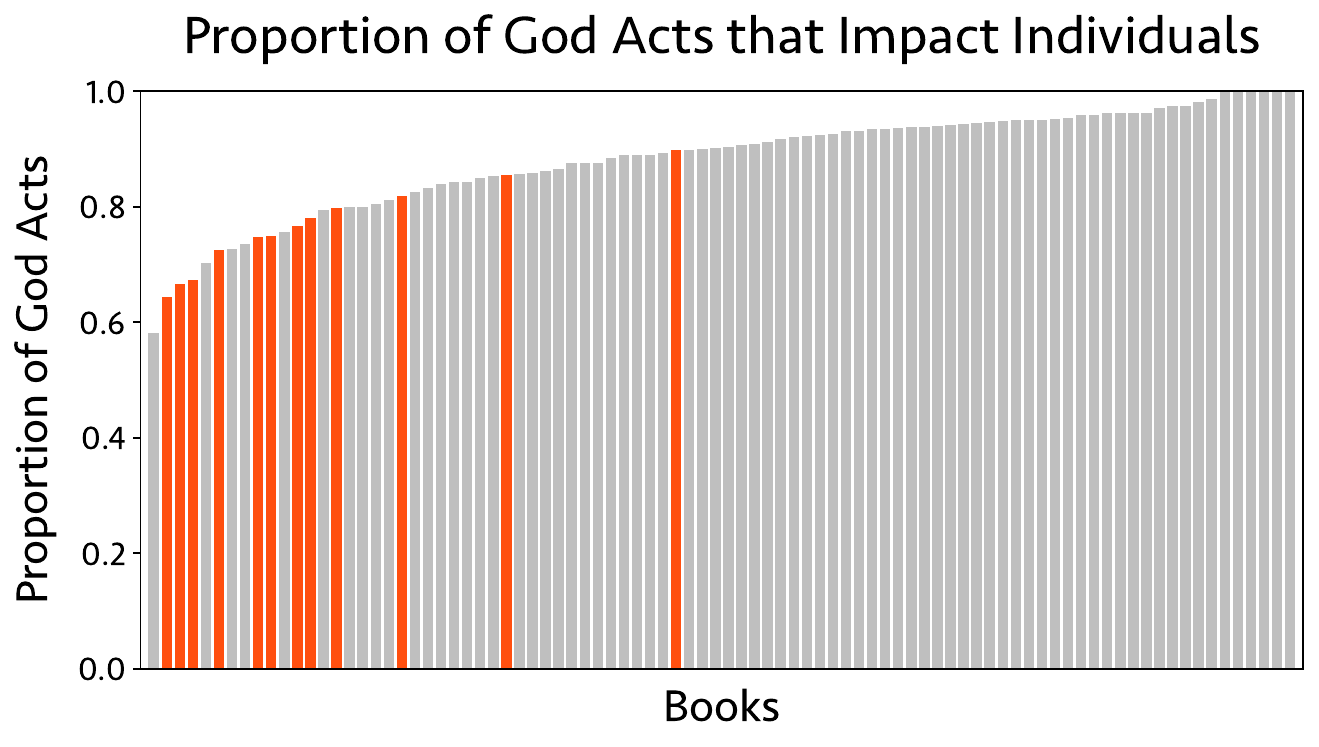}
    \caption{The proportion of passages labeled as containing acts of God that impact individuals. The \textit{Left Behind} novels, shown in orange, have some of the lowest proportions of divine acts affecting individuals.}
    \label{fig:propIndiv}
\end{figure*}

\begin{figure*}[t]
    \centering
    \includegraphics[width=.65\linewidth]{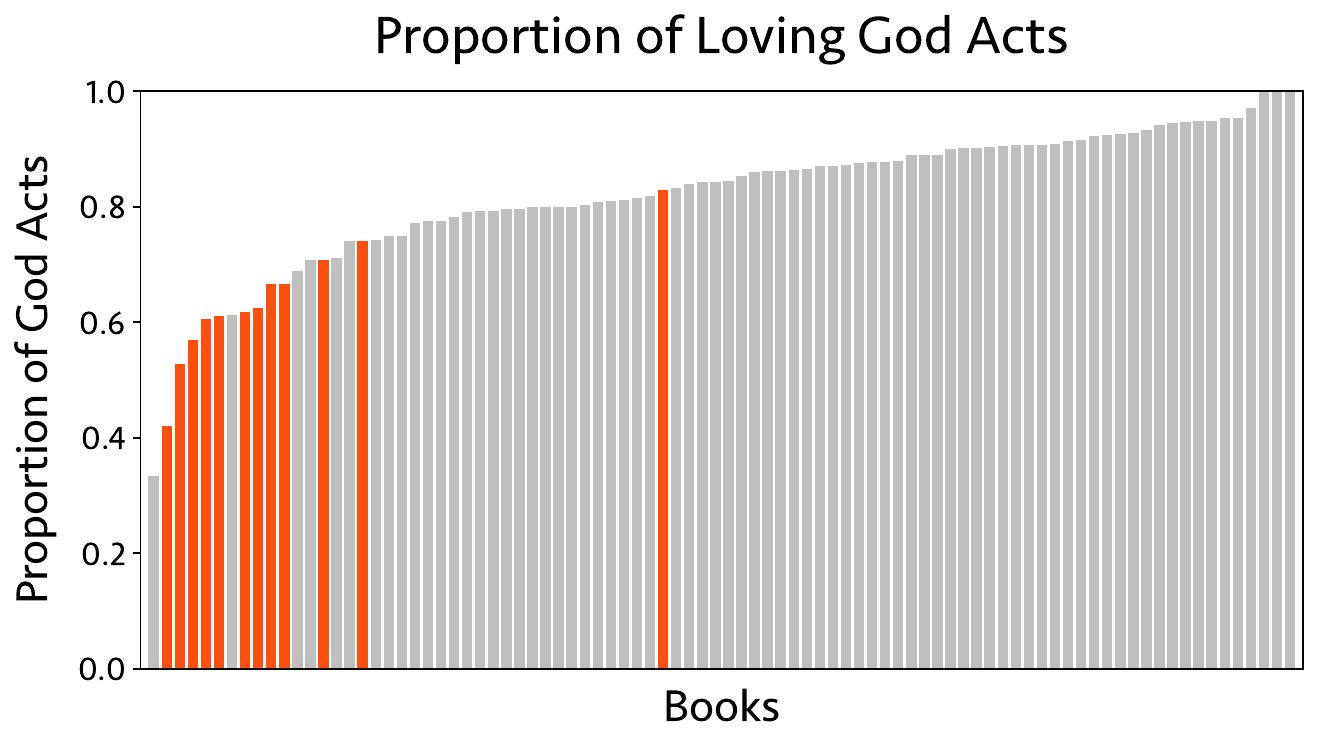}
    \caption{The proportion of passages labeled as containing acts of God that labeled as loving. The \textit{Left Behind} novels, shown in orange, have some of the lowest proportions of loving divine acts.}
    \label{fig:propLove}
\end{figure*}

We find that the overwhelming majority of divine acts in our corpus are loving acts that affect individuals; this trend is notable even taking into account that our classifiers tend to over-predict these labels. Of all the acts of God identified in our corpus, 86.8\% are labeled as affecting individuals and 13.2\% affect groups. 79.4\% of divine acts are classified as loving, 16.1\% as both loving and punishing, 3.6\% as punishing, and 0.9\% as neutral. Although the majority of acts directed at both individuals and groups are loving, a far greater proportion of acts impacting groups are punishing in nature (6.25\% vs.\ 3.22\%) or both loving and punishing (29.17\% vs.\ 14.16\%). This reveals a difference in how God's actions are characterized depending on who is affected. 

The frequency of divine acts impacting individuals is negatively correlated with Apocalyptic Faith (Pearson's R: -0.65, $p<10^{-10}$), which is intuitive given apocalyptic acts often affect groups. A negative correlation also exists with Glorifying God (Pearson's R: -0.58, $p<10^{-7}$), which is surprising since this topic closely aligns with conventional Evangelical individualistic expressions of faith. There is also some connection between the characterization of divine acts and faith. The frequency of loving acts is negatively correlated with Apocalyptic Faith (Pearson's R: -0.67, $p<10^{-11}$) and Glorifying God (Pearson's R: -0.58, $p<10^{-7}$) and the frequency of acts that are both loving and punishing is positively correlated with both these topics (0.72, $p<10^{-13}$ --- 0.65, $p<10^{-10}$). Combined, these trends demonstrate that apocalyptic themes and the glorification of God are associated with a greater frequency of punishing divine acts and acts of God which impact groups.\footnote{Correlations largely do not hold without \textit{Left Behind} books, largely due to the small number of group and punishing acts that appear outside of them.}

We again find that the \textit{Left Behind} novels are unique within our corpus. On average, a significantly smaller proportion of divine acts in these books are targeted towards individuals (76.03\% vs.\ 90.11\%, $p<10^{-7}$, Figure \ref{fig:propIndiv}), likely because they depict how God transforms the world in preparation for His Second Coming. We also see that a significantly smaller proportion of God's actions in the \textit{Left Behind} books are depicted as loving (63.24\% vs.\ 84.94\%, $p<10^{-10}$, Figure \ref{fig:propLove}), while a significantly larger proportion of acts are both loving and punishing (28.73\% vs.\ 11.31\%, $p<10^{-11}$). Moreover, nearly twice the proportion of actions are punishing on average (7.00\% vs.\ 2.81\%, $p<0.05$). These differences reinforce that the apocalyptic narrative of the \textit{Left Behind} series is not typical of Christian Fiction as a genre and the books should not be used as a proxy for the genre as a whole.
Finally, we find no significant evidence that male and female authors differ in the frequency with which they portray acts of God or how they characterize these acts. 


\section{Conclusion}

Our multidisciplinary work provides a combination of literary, cultural, and methodological insights.
Scholarship has paid little attention to Christian Fiction published after the conclusion of the \textit{Left Behind} series, instead focusing on the male-dominated cultural movements seeking to assert Evangelical Christianity through cultural power. In contrast, the Christian Fiction community has adopted a fundamentally different approach: one where Christian women, organized around faith and fiction, seek to transform the individual hearts and minds of Americans. This work begins the project of describing God's portrayal in Christian Fiction, and thereby adds to our understanding of Evangelical ministry and faith. Ultimately, we find that the \textit{Left Behind} series fundamentally differs from predominant Christian Fiction novels due to its depiction of a more active and punishing God while the authors who dominate Christian Fiction circles today largely write stories that model Evangelical femininity and depict a loving God.

From a methodological perspective, while it may be tempting --- or terrifying --- to think of new computational technologies as replacing scholarship with mere button pushing, we find that effectively using these methods requires an extensive, iterative process.
More established methods such as topic modeling produce results of a clearly limited and general character that both facilitate and require interpretation.
Newer prompt-based methods that both interpret text and produce human-readable outputs still require serious work before we understand what we want to ask of them and how we want to ask for it.
Additionally, smaller LMs that are computationally accessible still struggle to annotate highly subtle phenomena with extremely high precision.
However, combining the expertise of subject-matter experts and traditional qualitative research methods with contemporary LMs allows for us to begin to study these complex topics on previously unreachable scales, opening possibilities for many new avenues of research.

\section*{Acknowledgements}

We would like to thank Erin Boell, Carolyn Chui, Estefania DeJesus, Alexandra Gier, Georgia Lawrence, Sushmi Majumder, and Elina Natarajan for their substantial annotation contributions. In addition,  we thank Axel Bax, Federica Bologna, Anna Choi, Sil Hamilton, Kiara Liu, Landon Schnabel, Rosamond Thalken, Andrea Wang, Matthew Wilkens, and Shengqi Zhu for their thoughtful feedback. This
work was supported in part by the NEH project AI for Humanists and the Society for the Scientific Study of Religion (SSSR) 2025 Student Research Grant.

\printbibliography

\appendix

\section{Corpus Information}

\subsection{Book List} \label{appdx:corpus}

{\scriptsize
\setlength{\tabcolsep}{2pt}

\centering
\begin{filecontents*}{corpus.csv}
Title,Author,Author Gender,1st Book,Publisher,Date of Publication,CF Publisher ,Award Category,Award Status,Award Year,Goodreads # Ratings,Goodreads Rating,Genre
Arena,Karen Hancock,Female,Yes,Bethany House Publishers,2002,Yes,Allegory / Fantasy,Winner,2003,2684,4.14,Fantasy
Dark Horse,John Fischer,Male,No,Revell / Baker Publishing Group,1983,Yes,Allegory / Fantasy,Finalist,2004,73,3.79,Christian Fiction
True to You,Becky Wade,Female,No,Bethany House Publishers,2017,Yes,Book of the Year & Contemporary Romance,Winner,2018,5693,4.14,Christian Fiction
The Edge of Belonging,Amanda Cox,Female,Yes,Revell / Baker Publishing Group,2020,Yes,Book of the Year & First Novel,Winner,2021,2880,4.44,Christian Fiction
Whose Waves These Are,Amanda Dykes,Female,Yes,Bethany House Publishers,2019,Yes,Book of the Year & First Novel & General Fiction,Winner / Finalist / Winner,2020,4335,4.36,Historical Fiction
Burning Sky,Lori Benton,Female,Yes,WaterBrook Multnomah Publishing Group,2013,Yes,Book of the Year & First Novel & Historical,Winner,2014,1758,4.31,Historical Fiction
Long Way Gone,Charles Martin,Male,No,Thomas Nelson,2016,Yes,Book of the Year & General Fiction,Finalist / Winner,2017,17981,4.36,Fiction
Within These Walls of Sorrow,Amanda Barratt,Female,No,Kregel Publications,2023,Yes,Book of the Year & Historical,Winner,2023,686,4.68,Historical Fiction
Becoming Mrs. Lewis,Patti Callahan,Female,No,Thomas Nelson,2018,Yes,Book of the Year & Historical Romance,Winner,2019,47840,4.05,Historical Fiction
Thief of Glory,Sigmund Brouwer,Male,No,WaterBrook Multnomah Publishing Group,2014,Yes,Book of the Year & Historical Romance,Winner,2015,868,4.24,Historical Fiction
The Five Times I Met Myself,James L. Rubart,Male,No,Thomas Nelson,2015,Yes,Book of the Year & Visionary,Winner,2016,1474,4.12,Christian Fiction
A Garden to Keep,Jamie Langston Turner,Female,No,Bethany House Publishers,2001,Yes,Contemporary,Winner,2002,487,4.05,Christian Fiction
Bookends,Liz Curtis Higgs,Female,No,WaterBrook Multnomah Publishing Group,2000,Yes,Contemporary,Finalist,2001,1603,3.84,Christian Fiction
Borders of the Heart,Chris Fabry,Male,No,Tyndale House Publishers,2012,Yes,Contemporary,Finalist,2013,781,3.9,Fiction
Embrace Me,Lisa Samson,Female,No,Thomas Nelson,2007,Yes,Contemporary,Finalist,2009,788,3.92,Christian Fiction
Finding Alice,Melody Carlson,Female,No,WaterBrook Multnomah Publishing Group,2003,Yes,Contemporary,Finalist,2004,2045,4.04,Fiction
In High Places,Tom Morrisey,Male,No,Bethany House Publishers,2007,Yes,Contemporary,Finalist,2008,159,3.97,Christian Fiction
No Dark Valley,Jamie Langston Turner,Female,No,Bethany House Publishers,2004,Yes,Contemporary,Finalist,2005,474,3.91,Christian Fiction
Not a Sparrow Falls,Linda Nichols,Female,No,Bethany House Publishers,2002,Yes,Contemporary,Finalist,2003,2153,4.12,Christian Fiction
Quaker Summer,Lisa Samson,Female,No,Thomas Nelson,2007,Yes,Contemporary,Finalist,2008,1223,3.72,Christian Fiction
Straight Up,Lisa Samson,Female,No,WaterBrook Multnomah Publishing Group,2006,Yes,Contemporary,Finalist,2007,245,3.67,Christian Fiction
The Air We Breathe,Christa Parrish,Female,No,Bethany House Publishers,2012,Yes,Contemporary,Finalist,2013,694,3.8,Christian Fiction
The Living End,Lisa Samson,Female,No,WaterBrook Multnomah Publishing Group,2003,Yes,Contemporary,Finalist,2004,300,3.77,Christian Fiction
Winter Birds,Jamie Langston Turner,Female,No,Bethany House Publishers,2006,Yes,Contemporary,Winner,2007,583,3.85,Fiction
All That It Takes,Nicole Deese,Female,No,Bethany House Publishers,2022,Yes,Contemporary Romance,Finalist,2023,1963,4.45,Christian Fiction
All That Really Matters,Nicole Deese,Female,No,Bethany House Publishers,2021,Yes,Contemporary Romance,Winner,2022,3427,4.4,Romance
How Sweet It Is,Alice J. Wisler,Female,No,Bethany House Publishers,2009,Yes,Contemporary Romance,Finalist,2010,647,3.66,Christian Fiction
Larkspur Cove,Lisa Wingate,Female,No,Bethany House Publishers,2010,Yes,Contemporary Romance,Finalist,2012,7427,4.01,Christian Fiction
My Foolish Heart,Susan May Warren,Female,No,Tyndale House Publishers,2011,Yes,Contemporary Romance,Finalist,2012,2539,4.24,Christian Fiction
Remembered,Tamera Alexander,Female,No,Bethany House Publishers,2007,Yes,Contemporary Romance,Winner,2008,3796,4.3,Christian Fiction
Sworn to Protect,DiAnn Mills,Female,No,Tyndale House Publishers,2010,Yes,Contemporary Romance,Winner,2011,1432,4.17,Christian Fiction
The Healer,Dee Henderson,Female,No,WaterBrook Multnomah Publishing Group,2002,Yes,Contemporary Romance,Finalist,2003,17113,4.39,Christian Fiction
The Measure of a Lady,Deeanne Gist,Female,No,Bethany House Publishers,2006,Yes,Contemporary Romance,Winner,2007,12706,3.94,Christian Fiction
True Devotion,Dee Henderson,Female,No,WaterBrook Multnomah Publishing Group,2008,Yes,Contemporary Romance,Finalist,2001,14155,4.28,Christian Fiction
Dangerous Passage,Lisa Harris,Female,No,Revell / Baker Publishing Group,2013,Yes,Contemporary Romance & Suspense,Winner,2014,4096,4.03,Mystery
Summer Snow,Nicole Baart,Female,No,Tyndale House Publishers,2008,Yes,Contemporary Series,Finalist,2009,851,3.96,Christian Fiction
The Waiting,Suzanne Woods Fisher,Female,No,Revell / Baker Publishing Group,2010,Yes,Contemporary Series,Finalist,2011,2944,4.34,Amish
You Don't Know Me,Susan May Warren,Female,No,Tyndale House Publishers,2012,Yes,Contemporary Series,Winner,2013,1693,4.3,Christian Fiction
Freedom's Ring,Heidi Chiavaroli,Female,Yes,Tyndale House Publishers,2017,Yes,First Novel,Finalist,2018,798,4.29,Historical Fiction
Irish Meadows,Susan Anne Mason,Female,No,Bethany House Publishers,2015,Yes,First Novel,Finalist,2016,2496,4,Historical Fiction
Roots of Wood and Stone,Amanda Wen,Female,Yes,Kregel Publications,2021,Yes,First Novel,Finalist,2021,774,4.44,Christian Fiction
The Mending String,Cliff Coon,Male,Yes,Moody Publishers,2004,Yes,First Novel,Winner,2005,110,3.84,Fiction
A Cast of Stones,Patrick W. Carr,Male,No,Bethany House Publishers,2013,Yes,First Novel & Visionary,Finalist,2014,5530,4.28,Fantasy
Auralia's Colors,Jeffrey Overstreet,Male,No,WaterBrook Multnomah Publishing Group,2007,Yes,First Novel & Visionary,Finalist,2008,1868,3.65,Fantasy
Apocalypse Dawn,Mel Odom,Female,No,Tyndale House Publishers,2003,Yes,Futuristic,Finalist,2004,2050,4.15,Christian Fiction
Apollyon,Tim LaHaye & Jerry B. Jenkins,Male,No,Tyndale House Publishers,1999,Yes,Futuristic,Finalist,2000,34110,4,Christian Fiction
Armageddon,Tim LaHaye & Jerry B. Jenkins,Male,No,Tyndale House Publishers,2003,Yes,Futuristic,Finalist,2004,21823,4.05,Christian Fiction
Assassins,Tim LaHaye & Jerry B. Jenkins,Male,No,Tyndale House Publishers,1999,Yes,Futuristic,Finalist,2000,34017,4.01,Christian Fiction
Desecration,Tim LaHaye & Jerry B. Jenkins,Male,No,Tyndale House Publishers,2001,Yes,Futuristic,Finalist,2002,28060,4.03,Christian Fiction
Life After,Katie Ganshert,Female,No,WaterBrook Multnomah Publishing Group,2017,Yes,General Fiction,Winner,2018,2313,4.23,Christian Fiction
The Secret Keepers of Old Depot Grocery,Amanda Cox,Female,No,Revell / Baker Publishing Group,2021,Yes,General Fiction,Winner,2022,2955,4.27,Christian Fiction
All for a Story,Allison Pittman,Female,No,Tyndale House Publishers,2013,Yes,Historical,Finalist,2014,329,3.57,Christian Fiction
Fire by Night,Lynn Austin,Female,No,Bethany House Publishers,2003,Yes,Historical,Winner,2004,9792,4.46,Historical Fiction
Madman,Tracy Groot,Female,No,Tyndale House Publishers,2006,Yes,Historical,Winner,2007,686,4.11,Christian Fiction
Memories of Glass,Melanie Dobson,Female,No,Tyndale House Publishers,2019,Yes,Historical,Finalist,2020,2933,4.14,Historical Fiction
The Medallion,Cathy Gohlke,Female,No,Tyndale House Publishers,2019,Yes,Historical,Winner,2020,1466,4.54,Historical Fiction
The Noble Fugitive,T. Davis Bunn & Isabella Bunn,Both ,No,Bethany House Publishers,2005,Yes,Historical,Finalist,2006,746,4.05,Christian Fiction
The Swiss Courier,Tricia Gover & Mike Yorkey,Both ,No,Revell / Baker Publishing Group,2009,Yes,Historical,Finalist,2010,993,3.92,Historical Fiction
The White Rose Resists,Amanda Barratt,Female,No,Kregel Publications,2020,Yes,Historical,Winner,2021,696,4.44,Historical Fiction
A Defense of Honor,Kristi Ann Hunter,Female,No,Bethany House Publishers,2018,Yes,Historical Romance,Finalist,2019,3131,4.11,Historical Fiction
A Portrait of Loyalty,Roseanna M. White,Female,No,Bethany House Publishers,2020,Yes,Historical Romance,Winner,2021,2249,4.46,Historical Fiction
Calico Canyon,Mary Connealy,Female,No,Barbour Publishing,2008,Yes,Historical Romance,Finalist,2009,2532,4.28,Christian Fiction
Harvest of Gold,Tessa Afshar,Female,No,"RiverNorth, imprint of Moody Publishing",2013,Yes,Historical Romance,Winner,2014,4531,4.39,Christian Fiction
The Rose and the Thistle,Laura Frantz,Female,No,Revell / Baker Publishing Group,2023,Yes,Historical Romance,Winner,2023,2941,4.15,Historical Fiction
The Sound of Light,Sarah Sundin,Female,No,Revell / Baker Publishing Group,2023,Yes,Historical Romance,Finalist,2023,3692,4.33,Historical Fiction
Unashamed,Francine Rivers,Female,No,Tyndale House Publishers,2000,Yes,International Historical,Winner,2001,13710,4.29,Christian Fiction
By Reason of Insanity,Randy Singer,Male,No,Tyndale House Publishers,2008,Yes,Mystery / Suspense / Thriller,Finalist,2009,2365,3.88,Fiction
By Way of Deception,Amir Tsarfati & Steve Yohn,Male,No,Harvest House Publishers,2022,Yes,Mystery / Suspense / Thriller,Finalist,2023,978,4.56,Christian Fiction
Mind Games,Nancy Mehl,Female,No,Bethany House Publishers,2018,Yes,Mystery / Suspense / Thriller,Finalist,2019,5122,4.13,Christian Fiction
The Girl Behind the Red Rope,Ted Dekker & Rachelle Dekker,Both ,No,Revell / Baker Publishing Group,2019,Yes,Mystery / Suspense / Thriller,Winner,2020,5655,3.75,Christian Fiction
The Rook,Steven James,Male,No,Revell / Baker Publishing Group,2008,Yes,Mystery / Suspense / Thriller,Winner,2009,5918,4.28,Mystery
Candle in the Darkness,Lynn Austin,Female,No,Bethany House Publishers,2002,Yes,North American Historical,Winner,2003,17455,4.32,Historical Fiction
The Meeting Place,Janette Oke & T. Davis Bunn,Both ,No,Bethany House Publishers,1999,Yes,North American Historical,Winner,2000,5888,4.12,Christian Fiction
Dark Intercept,Brian Andrews & Jeffrey Wilson,Male,No,Tyndale House Publishers,2021,Yes,Speculative,Finalist,2022,2165,4.07,Thriller
Dark Justice,Brandilyn Collins,Female ,No,Broadman & Holman Publishers,2013,Yes,Suspense,Finalist,2014,875,4.25,Christian Fiction
The Last Plea Bargain,Randy Singer,Male,No,Tyndale House Publishers,2012,Yes,Suspense,Finalist,2013,1580,4.2,Christian Fiction
Shadow Hand,Anne Elisabeth Stengl,Female,No,Bethany House Publishers,2014,Yes,Visionary,Finalist,2015,666,4.32,Fantasy
Valley of the Shadow,Tom Pawlik,Male,No,Tyndale House Publishers,2009,Yes,Visionary,Finalist,2010,251,4.15,Christian Fiction
The Long Trail Home,Stephen Bly,Male,No,Broadman & Holman Publishers,2001,Yes,Western,Winner,2002,125,4.27,Westerns
"Motorcycles, Sushi, and One Strange Book",Nancy Rue,Female,No,Zondervan,2010,Yes,Young Adult,Winner,2011,567,3.88,Young Adult
Glorious Appearing,Tim LaHaye & Jerry B. Jenkins,Male,No,Tyndale House Publishers,2004,Yes,NA,NA,NA,,,
Left Behind,Tim LaHaye & Jerry B. Jenkins,Male,No,Tyndale House Publishers,1995,Yes,NA,NA,NA,,,
Nicolae,Tim LaHaye & Jerry B. Jenkins,Male,No,Tyndale House Publishers,1997,Yes,NA,NA,NA,,,
Soul Harvest,Tim LaHaye & Jerry B. Jenkins,Male,No,Tyndale House Publishers,1998,Yes,NA,NA,NA,,,
The Indwelling,Tim LaHaye & Jerry B. Jenkins,Male,No,Tyndale House Publishers,2000,Yes,NA,NA,NA,,,
The Mark,Tim LaHaye & Jerry B. Jenkins,Male,No,Tyndale House Publishers,2000,Yes,NA,NA,NA,,,
The Remnant,Tim LaHaye & Jerry B. Jenkins,Male,No,Tyndale House Publishers,2002,Yes,NA,NA,NA,,,
Tribulation Force,Tim LaHaye & Jerry B. Jenkins,Male,No,Tyndale House Publishers,1996,Yes,NA,NA,NA,,,
\end{filecontents*}
\csvreader[
    separator=comma,
    respect all,
    longtable=|
        >{\columncolor{gray!10}\centering\arraybackslash}m{2cm}|  
        >{\centering\arraybackslash}m{2cm}|                      
        >{\columncolor{gray!10}\centering\arraybackslash}m{1cm}|
        >{\centering\arraybackslash}m{2.5cm}|   
        >{\columncolor{gray!10}\centering\arraybackslash}m{1cm}|                      
        >{\centering\arraybackslash}m{2cm}|                        
        >{\columncolor{gray!10}\centering\arraybackslash}m{1.5cm}|   
        >{\centering\arraybackslash}m{1cm}|                        
,
     table head=\caption{The full subcorpus of Christian Fiction texts analyzed.}\label{tab:corpus}\\\hline
        \textbf{Title} & \textbf{Author} & \textbf{Gender} & \textbf{Publisher} & \textbf{Year} & \textbf{Award} & \textbf{Status} & \textbf{Award Year}  \\\hline\endfirsthead
        \hline \textbf{Title} & \textbf{Author} & \textbf{Gender} & \textbf{Publisher} & \textbf{Year} & \textbf{Award} & \textbf{Status} & \textbf{Award Year} \\\hline\endhead
        \hline \multicolumn{8}{r}{{}} \\\hline\endfoot
        \hline\endlastfoot,
    late after line=\tabularnewline\hline
]{corpus.csv}{}%
{\csvcoli & \csvcolii & \csvcoliii & \csvcolv & \csvcolvi & \csvcolviii & \csvcolix & \csvcolx}
}

\subsection{Books by Award Category} \label{appdx:corpusByCat}

{\scriptsize
\setlength{\tabcolsep}{2pt}

\centering
\begin{filecontents*}{corpus_count.csv}
Award Category,# Books
Contemporary,13
Contemporary Romance,12
Historical,10
Book of the Year,9
First Novel,9
Historical Romance,8
Mystery / Suspense / Thriller,5
Futuristic,5
Visionary,5
General Fiction,4
Suspense,3
Contemporary Series,3
Allegory / Fantasy,2
North American Historical,2
International Historical,1
Young Adult,1
Speculative,1
Western,1
\end{filecontents*}
\csvreader[
    separator=comma,
    respect all,
    longtable=|
        >{\columncolor{gray!10}\centering\arraybackslash}m{4cm}|  
        >{\centering\arraybackslash}m{2cm}|                      
,
     table head=\caption{The number of books in our corpus in each Award Category.}\label{tab:corpusByCat}\\\hline
        \textbf{Award Category} & \textbf{\# Books} \\\hline\endfirsthead
        \hline \textbf{Award Category} & \textbf{\# Books} \\\hline\endhead
        \hline \multicolumn{2}{r}{{}} \\\hline\endfoot
        \hline\endlastfoot,
    late after line=\tabularnewline\hline
]{corpus_count.csv}{}%
{\csvcoli & \csvcolii}
}

\subsection{Christy Award Categories by Author Gender} \label{appdx:catByGender}

{\scriptsize
\setlength{\tabcolsep}{2pt}

\centering
\begin{filecontents*}{christy_gender.csv}
Award Category,# Books,% Female Authors
Amplify Award,3,100.0
Lits,7,100.0
Contemporary Romance,79,98.73
Historical Romance,52,98.08
Short Form,27,92.59
Young Adult,54,85.19
Historical,67,85.07
Contemporary Series,33,84.85
Speculative,13,84.62
First Novel,75,76.0
North American Historical,12,75.0
Book of the Year,11,72.73
General Fiction,23,69.57
Contemporary,52,69.23
Suspense,35,57.14
Visionary,47,55.32
Allegory / Fantasy,6,50.0
Mystery / Suspense / Thriller,50,48.0
International Historical,12,41.67
Futuristic,15,13.33
Western,6,0.0
\end{filecontents*}
\csvreader[
    separator=comma,
    respect all,
    longtable=|
        >{\columncolor{gray!10}\centering\arraybackslash}m{4cm}|  
        >{\centering\arraybackslash}m{2cm}|                      
        >{\columncolor{gray!10}\centering\arraybackslash}m{2cm}|  
,
     table head=\caption{The number of books honored in each award category in the entire Christy Award corpus and the percentage of honorees in that category written only by female authors.}\label{tab:catByGender}\\\hline
        \textbf{Award Category} & \textbf{\# Books} & \textbf{\% Female Authors} \\\hline\endfirsthead
        \hline \textbf{Award Category} & \textbf{\# Books} & \textbf{\% Female Authors} \\\hline\endhead
        \hline \multicolumn{3}{r}{{}} \\\hline\endfoot
        \hline\endlastfoot,
    late after line=\tabularnewline\hline
]{christy_gender.csv}{}%
{\csvcoli & \csvcolii & \csvcoliii}
}

\section{Annotation Instructions and Prompts}

\subsection{\textit{Act of God} Instructions for Human Annotators} \label{appdx:humanActPrompt}

Is God acting or is an act of God described? -- Code: 1 = Yes, 2 = Maybe, 3 = No
\begin{enumerate}
    \item Yes: In the passage something is clearly ascribed to God.
    \begin{enumerate}
        \item NOTE: If there is an active Verb (including love), that’s an Act that we trust unless there is \textbf{narrative uncertainty}.
        \item NOTE: If God is described as one who does an active verb (God who provided / the One who gives me marching orders) this is a Yes, as an action is occurring or has occurred.
        \item NOTE: Descriptions and Quotes from the Bible that describe acts are acts of God. 
    \end{enumerate}
    \item Maybe: Something is insinuated to be due to God, but it is left ambiguous.
    \begin{enumerate}
        \item NOTE: The maybe category is designed for the natural coincidences (like the perfect sunsets), the narrative uncertainties (was that really God?), and mystical language that highly suggests that God is acting (something keeps her here).
        \item NOTE: In order for something to be ascribed to God (Yes), there must be \textbf{narrative certainty}. If the narrator or non-internally focalized text confirms or disputes the intervention being God’s, we trust that. If a character says something that is not questioned, we trust that it is legitimate. If we see a character ascribe action to God that is questioned by other characters or the narration, we don’t trust that God has acted. Ultimately, if the author introduces DOUBT about it being God, we DOUBT too. If there is no DOUBT in the passage that it is God’s act, then we trust it. 
        \item NOTE: If God is thanked in response to an action, it is implied the action might be attributed to God, so this is a Maybe.
    \end{enumerate}
    \item No: No mention is made in the passage of an action by God.
    \begin{enumerate}
        \item NOTE: Future tense (God will do…) is a NO. As are descriptors of God (God is kind / loving / powerful / righteous) and descriptors of individuals (God’s chosen one / God’s loved one).
    \end{enumerate}
\end{enumerate}

\subsection{Model Prompts for Characterizing Acts of God} \label{appdx:characPrompts}

\textbf{Prompt: Who does God affect?}
\begin{verbatim}
You will be given a description of an act of the Christian God in a novel 
passage. Decide who the Christian God is affecting in the passage.

Choose one of the following codes: 
- INDIVIDUAL: God affects one person.
- GROUP: God affects a group or community (e.g., a church, a town, a book 
club).

Please respond with:
- god_affect_explanation: Explain why you chose INDIVIDUAL or GROUP
- god_affect: INDIVIDUAL or GROUP

<text>
[INSERT TEXT HERE]
</text>
\end{verbatim}
\: \\
\textbf{Prompt: What is the intent of God's act?}
\begin{verbatim}
You will be given a description of an act of the Christian God in a novel 
passage. Decide what kind of action it is.

Choose one of the following codes: 
- LOVING: God's action is kind (for example it invovles mercy, love, 
forgiveness, or help).
- PUNISHING: God's action is meant to punish or judge (for example it involves 
anger, vengeance, violence, or judgment).
- BOTH: God's action has elements of both love and punishment.
- NEUTRAL: God's action is neutral or ambiguous. Avoid using this label when 
possible.

Please respond with:
- god_impact_explanation: Explain why you chose LOVING, PUNISHING, BOTH, or 
NEUTRAL
- god_impact: LOVING, PUNISHING, BOTH, or NEUTRAL

<text>
[INSERT TEXT HERE]
</text>
\end{verbatim}

\section{Topic Modeling Results}

\subsection{Topic Labels and Keywords} \label{appdx:topicLabels}

{\scriptsize
\setlength{\tabcolsep}{2pt}

\centering
\begin{filecontents*}{topic_labels_words.tsv}
Topic	Label	Words
0	Measuring Time	minutes, three, five, ten, four, six, later, hundred, twenty, watch
1	Kitchen + Dining	table, food, eat, kitchen, bread, plate, bowl, set, dinner, bite
2	Joyful Affection	love, heart, smile, eyes, kiss, loved, joy, beautiful, together, kissed
3	Goodbyes	won't, leave, you'll, stay, we'll, better, head, else, wouldn't, you've
4	Familial Relationships	family, father, children, mother, parents, son, daughter, child, brother, name
5	Reading + Writing	read, book, reading, story, books, words, page, writing, bible, write
6	Groups of People	men, women, others, each, line, three, group, camp, guards, guard
7	Digital Communication	phone, call, number, cell, voice, called, message, name, calling, rang
8	Familial Emotions + Hardships	mom, dad, mother, okay, father, she's, sorry, happened, love, kids
9	Body + Intimacy	eyes, air, voice, head, skin, body, chest, hand, open, arms
10	Young Adult Life	pretty, kind, she's, nice, girl, probably, lot, bad, guy, real
11	Financial Assets	money, pay, business, buy, job, store, dollars, house, paid, sell
12	Perception of Self + Others	eyes, face, seen, herself, seemed, such, smile, almost, caught, eye
13	Confessional Prayer	done, must, truth, myself, love, believe, father, understand, feel, god
14	Familial Events	father, mother, aunt, father's, daddy, mama, mother's, uncle, family, herself
15	Body Language	head, eyes, hand, shook, hands, leaned, face, chair, nodded, glanced
16	Sleeping	bed, night, sleep, morning, room, asleep, sleeping, hours, slept, awake
17	Urban Movement	street, building, walked, past, walk, walking, corner, guard, steps, crowd
18	Affectionate Gestures	smile, hand, miss, smiled, thank, gave, young, nodded, meet, please
19	Education	school, year, high, college, class, kids, during, teacher, ago, summer
20	State Conflict + WW2	men, war, soldiers, army, general, city, military, battle, news, german
21	Sky Appearance	light, its, dark, white, sky, black, sun, above, red, darkness
22	Nations + International Community	world, community, meeting, news, believe, each, states, united, everyone, become
23	Violence	head, blood, ground, feet, body, face, gun, arm, hands, dead
24	Stationary	paper, letter, read, letters, desk, envelope, written, note, write, name
25	Sports	game, play, ball, team, played, playing, games, win, second, field
26	Friendship	days, each, friends, friend, times, seemed, since, often, part, spent
27	Motherhood + Infancy	mrs, baby, child, mother, husband, girl, ma'am, young, women, she's
28	Rural Nature	trees, tree, house, its, grass, ground, woods, along, leaves, path
29	Formal Clothing	dress, wearing, white, shirt, black, shoes, wore, coat, clothes, blue
30	States of Being	someone, seemed, hadn't, finally, looking, else, waited, started, room, seen
31	Casual Conversation	okay, guy, head, guess, better, lot, pretty, you've, stuff, doing
32	Home	house, rachel, night, kitchen, dinner, table, morning, today, later, already
33	Doubtful + Unsure	hadn't, wouldn't, trying, everything, part, needed, myself, thinking, happened, else
34	Cleaning the Home	water, clean, clothes, kitchen, sink, hot, hands, bathroom, towel, wash
35	Violent Crime	death, dead, killed, kill, believe, prison, murder, story, tried, justice
36	House Features	room, floor, house, open, wall, window, table, kitchen, stairs, hall
37	Technology	computer, screen, video, name, information, show, camera, miles, second, list
38	Physical Features	hair, eyes, face, dark, brown, white, skin, black, tall, gray
39	Aircraft	plane, area, air, pilot, radio, fly, hours, ground, clear, phone
40	Youthful Playing	boy, boys, son, laughed, girl, laughing, laugh, laughter, name, mouth
41	Law Enforcement	police, case, officer, information, someone, agent, evidence, needed, involved, working
42	Photographs	box, bag, picture, ring, opened, open, small, set, photo, counter
43	Glorious God	god, lord, jesus, pray, bible, god's, love, faith, christ, prayer
44	Intimate Touch	hand, eyes, face, hands, tears, arm, head, held, pulled, reached
45	Physical Nervousness	eyes, voice, face, breath, throat, hands, chest, head, heart, mouth
46	Grief + Emotional Turmoil	heart, fear, world, words, lost, pain, voice, gone, death, eyes
47	Listening	says, looks, hand, she's, hands, head, takes, eyes, turns, comes
48	Apocalyptic Faith	god, shall, world, great, israel, crowd, earth, jerusalem, michael, power
49	Human Movement	feet, open, hand, foot, onto, under, pulled, steps, reached, top
50	Passage of Time	morning, night, days, hadn't, since, week, weeks, afternoon, three, they'd
51	Ocean + Shore	water, boat, river, lake, sea, bridge, shore, its, fish, cold
52	Horses	horse, wind, rain, horses, sky, far, its, ground, clouds, feet
53	Medical Pain	doctor, hospital, room, pain, blood, medical, nurse, body, leg, bed
54	Rural Towns + Travel	city, town, train, north, street, west, south, road, streets, river
55	Beverages + Drinking	coffee, cup, tea, drink, table, glass, water, bottle, set, sip
56	Communication	voice, words, hear, answer, question, spoke, speak, word, ask, understand
57	Motor Vehicles	car, road, seat, truck, pulled, drive, window, parking, drove, lot
58	Questioning	such, question, doubt, already, point, certain, between, fact, given, best
59	Royalty	king, upon, such, must, lord, great, men, speak, shall, far
60	Childhood Pets	girl, children, child, boy, dog, cat, name, small, eyes, sister
61	Fire	fire, smoke, wood, flames, burning, set, heat, air, burned, pieces
62	Holidays (Christmas)	christmas, red, white, gift, art, paint, color, flowers, beautiful, painting
63	Congregation Worship	church, music, song, pastor, sunday, singing, service, sing, dance, sang
64	Passage of Time	since, perhaps, days, leaving, such, certain, leave, window, its, evening
\end{filecontents*}
\csvreader[
    separator=tab,
    respect all,
    longtable=|
        >{\columncolor{gray!10}\centering\arraybackslash}m{1cm}|  
        >{\centering\arraybackslash}m{3cm}|                      
        >{\columncolor{gray!10}\centering\arraybackslash}m{7cm}|  
,
     table head=\caption{Each of the 65 topics with human-annotated labels and the top 10 keywords.}\label{tab:topicLabels}\\\hline
        \textbf{Topic} & \textbf{Label} & \textbf{Words} \\\hline\endfirsthead
        \hline \textbf{Topic} & \textbf{Label} & \textbf{Words} \\\hline\endhead
        \hline \multicolumn{3}{r}{{}} \\\hline\endfoot
        \hline\endlastfoot,
    late after line=\tabularnewline\hline
]{topic_labels_words.tsv}{}%
{\csvcoli & \csvcolii & \csvcoliii}
}

\subsection{Topics Ranked by Prominence} \label{appdx:topicProm}

{\scriptsize
\setlength{\tabcolsep}{2pt}

\centering
\begin{filecontents*}{topic_prom.csv}
Topic,Label,Average Novel Prominence
23,Violence,3.86
58,Questioning,3.48
33,Doubtful + Unsure,3.44
13,Confessional Prayer,3.28
31,Casual Conversation,3.24
45,Physical Nervousness,3.2
3,Goodbyes,2.96
44,Intimate Touch,2.42
15,Body Language,2.42
46,Grief + Emotional Turmoil,2.25
18,Affectionate Gestures,2.21
30,States of Being,2.16
49,Human Movement,2.06
36,House Features,2.02
4,Familial Relationships,1.9
43,Glorious God,1.81
47,Listening,1.8
59,Royalty,1.75
52,Horses,1.7
26,Friendship,1.68
9,Body + Intimacy,1.64
28,Rural Nature,1.64
21,Sky Appearance,1.63
2,Joyful Affection,1.6
12,Perception of Self + Others,1.59
1,Kitchen + Dining,1.57
50,Passage of Time,1.55
22,Nations + International Community,1.53
39,Aircraft,1.51
0,Measuring Time,1.51
56,Communication,1.45
10,Young Adult Life,1.44
38,Physical Features,1.41
16,Sleeping,1.41
17,Urban Movement,1.35
7,Digital Communication,1.26
19,Education,1.25
57,Motor Vehicles,1.22
8,Familial Emotions + Hardships,1.22
20,State Conflict + WW2,1.19
41,Law Enforcement,1.18
29,Formal Clothing,1.17
11,Financial Assets,1.15
48,Apocalyptic Faith,1.09
6,Groups of People,1.07
35,Violent Crime,1.06
64,Passage of Time,1.01
32,Home,0.97
5,Reading + Writing,0.92
24,Stationary,0.86
54,Rural Towns + Travel,0.85
62,Holidays (Christmas),0.85
37,Technology,0.83
51,Ocean + Shore,0.79
42,Photographs,0.78
61,Fire,0.75
60,Childhood Pets,0.72
63,Congregation Worship,0.72
34,Cleaning the Home,0.71
55,Beverages + Drinking,0.7
40,Youthful Playing,0.68
53,Medical Pain,0.66
25,Sports,0.64
14,Familial Events,0.62
27,Motherhood + Infancy,0.61
\end{filecontents*}
\csvreader[
    separator=comma,
    respect all,
    longtable=|
        >{\columncolor{gray!10}\centering\arraybackslash}m{1cm}|  
        >{\centering\arraybackslash}m{3cm}|                      
        >{\columncolor{gray!10}\centering\arraybackslash}m{3.5cm}|  
,
     table head=\caption{The topics ranked by average novel prominence.}\label{tab:topicProps}\\\hline
        \textbf{Topic} & \textbf{Label} & \textbf{Average Novel Prominence} \\\hline\endfirsthead
        \hline \textbf{Topic} & \textbf{Label} & \textbf{Average Novel Prominence} \\\hline\endhead
        \hline \multicolumn{3}{r}{{}} \\\hline\endfoot
        \hline\endlastfoot,
    late after line=\tabularnewline\hline
]{topic_prom.csv}{}%
{\csvcoli & \csvcolii & \csvcoliii}
}

\end{document}